# Probabilistic Models for Query Approximation with Large Sparse Binary Data Sets.

**Dmitry Pavlov**
Information and Computer Science
University of California, Irvine
CA 92697-3425
`pavlovd@ics.uci.edu`

**Heikki Mannila**
Nokia Research Center
P.O. Box 407
FIN-00045 Nokia Group, Finland
`Heikki.Mannila@nokia.com`

**Padhraic Smyth**
Information and Computer Science
University of California, Irvine
CA 92697-3425
`smyth@ics.uci.edu`

## Abstract

Large sparse sets of binary transaction data with millions of records and thousands of attributes occur in various domains: customers purchasing products, users visiting web pages, and documents containing words are just three typical examples. Real-time query selectivity estimation (the problem of estimating the number of rows in the data satisfying a given predicate) is an important practical problem for such databases.

We investigate the application of probabilistic models to this problem. In particular, we study a Markov random field (MRF) approach based on frequent sets and maximum entropy, and compare it to the independence model and the Chow-Liu tree model. We find that the MRF model provides substantially more accurate probability estimates than the other methods but is more expensive from a computational and memory viewpoint. To alleviate the computational requirements we show how one can apply bucket elimination and clique tree approaches to take advantage of structure in the models and in the queries. We provide experimental results on two large real-world transaction datasets.

## 1 Introduction

Massive datasets containing huge numbers of records have recently become an object of increasing interest among both the businesses who routinely collect such data and data miners who try to find regularities in them. One class of such datasets is transaction data which is typically binary in nature. This class is characterized by high *sparseness*, i.e., there may be hundreds and thousands of binary attributes but a particular record may only have a few of them set to 1.

Examples include retail transaction data and Web log data, where each row is a transaction or session and each column represents a product or Web page.

Data-owners typically have a lot of questions about their data. For instance, it may be of interest to know how often the pages $W_1$ and $W_2$ but not $W_3$ were requested together. Such types of questions about the data can be formalized as Boolean queries on arbitrary subsets of attributes. The problem then is to find the frequency of rows in the dataset that satisfy query $Q$, or, equivalently the probability of the query $P(Q)$ with respect to the empirical probability distribution defined by the data.

Any Boolean query can be answered using a single scan through the dataset. While this approach has linear complexity and works well for the small datasets, it becomes infeasible for real time queries on huge datasets which do not reside in main memory. We would thus like to have an approximate algorithm that would allow us to trade accuracy in the estimate of $P(Q)$ with the time taken to calculate it. This paper studies different probabilistic models for approximating $P(Q)$.

A simple model-based approach is to calculate the marginal frequencies of all attributes and use an independence model for $P(Q)$ (often the method of choice in commercial relational database systems). The independence model is easy to learn, has low time and space requirements but as we shall see below is fairly inaccurate. [Mannila et al., 1999] introduced the idea of using an MRF model based on frequent itemsets and a maximum entropy (*maxent*) approach for query approximation. The motivation comes from the fact that there exist many efficient data mining algorithms for extracting frequent itemsets from massive data sets, and maxent principle can be used to combine these itemsets to form a coherent probabilistic MRF model.

In this paper we show that the itemsets and the maxent distribution define a *Markov random field (MRF)* by the fundamental MRF theorem [Hammersley and Clifford, 1972]. We also improve the standard iterative scaling (IS) algorithm for learning parameters of the MRF models by showing how one can apply bucket elimination and clique tree approaches to take advantage of structure in the models and in the queries. We show that depending on the number of itemsets used as an input to the maxent models their prediction accuracy averages within 0.1-1% of the true count.

We also investigate an approach based on tree-structured belief networks [Chow and Liu, 1968] that fills in the gap between the computationally inexpensive (but not very accurate) independence model and the relatively expensive maxent solution. We provide experimental results on the performance of all the methods on the real data. We compare models in terms of the accuracy of the approximation, and the time and amount of information that the model requires.

The rest of this paper is organized as follows. In Section 2 we introduce notation and give a formal statement of the estimation problem. Section 3 gives precise definitions of the models that we apply to the query selectivity estimation problem and the methodology we use to compare them. Section 4 presents the empirical results and in Section 5 we draw conclusions and present some extensions.

## 2 Statement of the Problem and Notation

Let $R = \{A_1, \ldots, A_k\}$ be a table header with $k$ 0/1 valued attributes (variables) and $r$ be a table of $n$ rows over header $R$. We assume that $k \ll n$, and that the data are sparse, as we discussed above. A row of the table $r$ satisfies a conjunctive query $Q$ iff the corresponding attributes in the query and in the row have equal values. We are interested in finding the number of rows in the table $r$ satisfying a given conjunctive query $Q$ defined on a subset of its attributes. We can view this *query selectivity estimation* problem in a probabilistic light and pose the problem as estimating the true frequency of $Q$ in the table $r$ using an approximate (and presumably much smaller and more efficient) probability model $P_M$.

The time involved in using a probability model $P_M$ is divided into the *offline cost* $T_P$, i.e., the time needed for building the model, and the *online cost* $t_p(Q)$ needed to give the approximate answer to query $Q$ using the model $P_M$. We use $S_P$ to denote the amount of space (memory) needed to store the model $P_M$.

For a given class of queries, let $\pi(Q)$ denote the probability that the query $Q$ is issued. We assume that this distribution is known, but in principle we could learn $\pi(Q)$ for a population or individuals. By $e_P(Q)$ we denote the error in answering the query $Q$, i.e., the difference between the true count $C_t(Q)$ and the count estimated from the model $P_M$. We are interested in the expectation of the relative error with respect to the underlying query distribution $E_\pi[|e_P(Q)|/C_t(Q)]$. We use the empirical relative error defined as

$$\hat{E} = \frac{1}{N_{Q's}} \sum_{j=1}^{N_{Queries}} \frac{|e_P(Q_j)|}{C_t(Q_j)}, \quad (1)$$

where $N_{Q's}$ is the number of random query drawings from $\pi(Q)$ and $C_t(Q_j)$ is the true count of the query $Q_j$.

## 3 Models

### 3.1 Full data and Independence model

There are a wide variety of options in choosing the model $P_M$. One extreme is to store the entire dataset so that for each record we will only keep a list of columns that have 1's in them. We will have 100% accurate estimates, but for most of the real-life datasets this approach will incur inordinately large memory requirements, namely $S_P = O(c \sum_{i=1}^{n} N_{1's}(i))$, where $c$ is the prespecified number of bits required to store a number to some fixed precision and $N_{1's}(i)$ is the number of positively initialized attributes for record $i$.

The other extreme is also easy to describe—*the independence model*. Since the data are binary-valued, we only have to store one count per attribute. The probability of a conjunctive query is approximated by the product of the probabilities of the single attribute-value combinations occurring in the query. Obviously, $S_P$ is small in this method, namely $O(kc)$ bits. The preprocessing can be done by a single scan through the data, and the online cost consists of $n_Q$ multiplications, where $n_Q$ is the number of conjuncts in the query $Q$. However, as we shall see later, the quality of the approximations produced by the independence method can be relatively poor.

### 3.2 Model Based on the Multivariate Tree Distribution

This model [Chow and Liu, 1968] assumes that there are only pairwise dependencies between the variables and that the dependency graph on the attributes is a tree. To fit a distribution with a tree it is sufficient to know the pairwise marginals of all the variables. The algorithm consists of three steps, namely, computing the pairwise marginals of the attributes, computing the mutual information between the attributes and, finally, applying Kruskal's algorithm to find the minimum spanning tree of the full graph whose nodes are the attributes and the weights on the edges are the mutual informations. The dominating term in the overall offline time complexity will be $O(k^2 n)$ due to the computation of the marginals. The memory requirements for the algorithm is $O(k^2 c)$. Once the tree is learned, we can use a standard belief propagation algorithm [Pearl, 1988] to get the answer to a particular conjunctive query $Q$ in time linear in $n_Q$.

### 3.3 Maximum Entropy MRF Model

An *itemset* associated with the binary table $r$ with the header $R$ is defined to be either a single positively initialized attribute or a conjunction of the mutually exclusive positively initialized attributes from $R$. We will call an itemset *T-frequent* if its count in the table $r$ is at least $T$ where $T$ is some predefined non-negative threshold.

There exist efficient algorithms to compute all the itemsets from large binary tables ( e.g., [Mannila et al., 1994, Agrawal and Srikant, 1994]). In practice the running time of these algorithms is linear in both the size of the table and the number of frequent itemsets provided that the data are sparse. Thus, by computing itemsets we won't typically incur a high preprocessing cost.

The maximum entropy approach makes use of the $T$-frequent itemsets and the associated frequency counts treating them as constraints on the query distribution. Indeed, each pair of (a) an itemset and (b) its associated frequency count can be viewed as a value of the marginal distribution on the query variables when they all are positively initialized.

Consider an arbitrary conjunctive query $Q$ on variables $x_Q = \{q_1, \ldots, q_{n_Q}\}$. Forcing the estimate $P_M$ to be consistent with the $T$-frequent itemsets for some $T > 0$ restricts $P_M$ to a constrained set $\mathcal{P}$ of probability distributions within the general $n_Q$-dimensional simplex containing all possible distributions defined on $n_Q$ variables. Information about frequencies of the $T$-frequent itemsets for some $T > 0$ in general underconstrains the target distribution and we will need an additional criterion to pick a unique estimate $P_M(Q)$ from the set $\mathcal{P}$ of all plausible ones. The maximum entropy principle provides such a criterion. It essentially instructs one to select a distribution that is as uninformed as possible, i.e., makes the fewest possible commitments about anything the constraints do not specify. Given maximum entropy as a preference criterion, we face a constrained optimization problem of finding $P_M(x_Q) = \arg\max_{P \in \mathcal{P}} H(P)$, where $H(P)$ is the entropy of the distribution $P$. If the constraints are consistent (which is clearly the case with itemset-based constraints) one can show that the target distribution will exist, be unique [Berger et al., 1996, Pietra et al., 1997] and can be found in an iterative fashion using an algorithm known as iterative scaling (IS) [Darroch and Ratcliff, 1972, Csiszár and Tusnády, 1984]. Note that we are estimating the full joint distribution on variables $x_Q$, not only the probability of the specific instantiation of variables in the given query $Q$.

In order to get a probability estimate $P_M(x_Q)$ we only retain itemsets whose variables are subsets of $x_Q$, i.e., the joint on $x_Q$ is estimated in real-time when a query is posed to the system. Enforcing the $j$-th constraint $c_j$ can be performed by just summing out from $P_M(x_Q)$ all the variables not participating in the $j$-th itemset (the variables in the constraint $c_j$ should be kept fixed to 1), and requiring that the result of this sum equals the true count $f_j$ of the $j$-th itemset in the table. Constraint $c_j$ will thus look like:

$$\sum_{x_Q \in \{0,1\}^{n_Q}} P_M(x_Q) G(A_1^j = 1, \ldots, A_{n_j}^j = 1) = f_j \quad (2)$$

where $G(.)$ is the indicator function.

Whenever we say that initialized query variables $x_Q$ *satisfy* a given constraint $c_j$, we shall mean that variables $x_Q$ agree in their values with all the variables $c_j$. It can be shown (see, e.g., [Jelinek, 1998]) that the maxent distribution will have a special product form

$$P_M(x_Q) = \mu_0 \prod_{j=1}^{N} \mu_j^{G(x_Q \ satisfies \ c_j)} \quad (3)$$

Once the constants $\mu_j$ are estimated, Equation 3 can be used to evaluate any query on the variables in $x_Q$, and $Q$ in particular. The product form of the target distribution will contain exactly one factor corresponding to each of the constraints. Factor $\mu_0$ is a normalization constant whose value is found from the condition

$$\sum_{x_Q \in \{0,1\}^{n_Q}} P_M(x_Q) = 1 \quad (4)$$

The general problem is thus reduced to the problem of finding a set of numbers $\mu_j$ from Equations 2 and 4. The IS algorithm is well known in the statistical literature as an iterative technique which converges to the maxent solution for problems of this general form (see, e.g., [Jelinek, 1998]). A high-level outline of the most computationally efficient version of the algorithm [Jelinek, 1998] is as follows.

1. Choose an initial approximation to $P_M(x_Q)$
2. While (Not all Constraints are Satisfied)
    For (j varying over all constraints)
        Update $\mu_0$;

   Update $\mu_j$;
  End;
 EndWhile;
3. Output the constants $\mu_j$

The update rules for parameter $\mu_j^t$ corresponding to the constraint $c_j$ at iteration $t$ are:

$$\mu_0^{t+1} = \mu_0^t \frac{1-f_j}{1-S_j^t} \quad (5)$$

$$\mu_j^{t+1} = \mu_j^t \frac{f_j(1-S_j^t)}{S_j^t(1-f_j)}, \quad (6)$$

where $S_j^t$ is defined as:

$$S_j^t = \sum_{x_Q \; satisfying \; c_j} P_M^t(x_Q) \quad (7)$$

Equation 7 essentially calculates the constraints as in Equation 2 but using the current estimate $P_M^t$ which in turn is defined by the estimates of the $\mu_j^t$'s via Equation 3. Since the distribution $P_M^t$ may not necessarily meet the $c_j$-th constraint, equations 5 and 6 update the terms $\mu_0$ and $\mu_j$ to enforce it. The algorithm proceeds in a round-robin fashion to the next constraint at each iteration getting closer to satisfying *all* of them.

Convergence of the algorithm can be determined by various means. In the case of the query selectivity estimation problem, we are interested only in one cell of the distribution on the query variables corresponding to a particular query $Q$. We monitor this particular cell and terminate the algorithm when

$$|P_M^t(Q) - P_M^{t-1}(Q)| < \varepsilon P_M^{t-1}(Q). \quad (8)$$

$\varepsilon$ is one of the free parameters of the algorithm, in the experiments we chose $\varepsilon = 10^{-4}$. It usually takes 10-15 iterations for the algorithm to meet such a convergence criterion.

Preprocessing for the maxent model consists of finding the itemsets for the entire dataset given a specified threshold $T$. Suppose that we have selected $N$ itemsets on the sets of variables $I_j = \{A_1^j, ..., A_{n_j}^j\}$ and we know their frequencies $f_j$ in the table $r$. Then the memory cost is $O(c(\sum_{k=1}^N n_k + N))$. The first term in this estimate corresponds to storing the attributes that are set to 1 in each of the itemsets, and the second term to storing the counts of the itemsets in the table $r$.

The main computation in the IS algorithm occurs in summing out the distribution $P_M^t(x_Q)$ according to Equation 7. The total number of summands in Equation 7 is $2^{n_Q-n_j}$. Each summand will have a product form and will contain at most $N$ factors. Thus, the overall time complexity of performing the summation in Equation 7 once for all the factors $\mu$ is $\sum_{j=1}^N \alpha_j 2^{\alpha_j}$, where $\alpha_j = n_Q - n_j$. The last estimate is obviously upper-bounded by $O(Nn_Q 2^{n_Q})$. Note that this is independent of the size of the original dataset. Although the exponential time complexity in the size of the query makes the method prohibitive for large query sizes, it is still feasible to use it for queries of length 8 or so in practice. The IS algorithm in its formulation above has linear memory complexity in $n_Q$ since the summation in Equation 7 can be performed using backtracking.

We note that the maxent distribution in Equation 3 is an MRF model. Suppose, that for a given query we have selected all itemsets that only mention query variables. These itemsets define an undirected *graphical model* $H$ on the query variables with an edge connecting the two nodes iff the corresponding variables are mentioned in some itemset. Each node $v$ in $H$ will naturally have a neighborhood — the set of all nodes in $H$ incident to $v$. Finally, from the product form of the maxent distribution in Equation 3, and the fact that each factor corresponds to an itemset, it follows that maxent distribution can be viewed as a product of exponential functions on the cliques of the graph $H$. Hence, by the fundamental MRF theorem [Hammersley and Clifford, 1972] maxent distribution defines a MRF with respect to the graph $H$. In the next subsection we discuss how one can speed-up the IS algorithm by trading memory for time based on the structure of the itemsets used to constrain the distribution.

### 3.4 Trading Memory for Time in Iterative Scaling

The strategies that we propose for reducing the computational complexity of the IS algorithm employ the structure in the queries and in the itemsets.

Below we show that in general the time complexity of IS can be reduced to being exponential in the *induced width* $w^*$ of the graph $H$. The notion of induced width is closely related to the size of cliques of the graph $H$ and can be thought of being equal to the size of the largest clique after the graph has been triangulated. Finding the ordering of the variables with width equal to $w^*$ is NP-hard in general. In our experiments we used the heuristic of maximum cardinality ordering.

#### 3.4.1 Bucket Elimination

*Bucket elimination* [Dechter, 1996] is essentially a smart way to do bookkeeping as one goes along the updates of factors in the representation of the maxent distribution. The idea here is to use the distributive law (see, e.g., [Aji and McEliece, 1997]) in Equation 7.

Consider a simple example. We issue a query on six binary attributes $A_1, \ldots, A_6$. There are frequent itemsets corresponding to each single attribute and the following frequent itemsets of size 2: $\{A_2 = 1, A_3 = 1\}$, $\{A_3 = 1, A_4 = 1\}$, $\{A_4 = 1, A_6 = 1\}$, $\{A_3 = 1, A_5 = 1\}$ and $\{A_5 = 1, A_6 = 1\}$. The graphical model $H$ in Figure 1 shows the interactions between the attributes in this example.

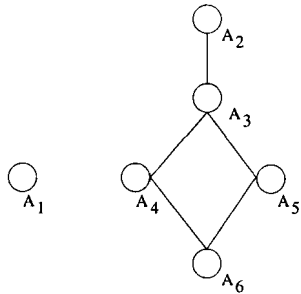

Figure 1: Graphical model for an example problem

The maxent distribution according to Equation 3 will have the following form

$$P = \mu_0 \prod_{i=1}^{6} \mu_i^{G(A_i=1)} \prod_{i,j} \mu_{ij}^{G(A_i=A_j=1)} G(\exists edge(i,j) \text{ in } H)$$

Suppose, that on the current iteration we are updating $\mu_{56}$ corresponding to itemset $\{A_5 = 1, A_6 = 1\}$. According to our update rule in Equation 7, we need to fix attributes $A_5$ and $A_6$ to 1 and sum out the rest of the attributes in $P^t(A_1, \ldots, A_6)$. It is easy to see that brute force summation over all the values of $A_1, \ldots, A_4$ will involve computing 16 terms, each having a product form. The bucket elimination algorithm will produce exactly the same result but will do it more efficiently—the number of terms to evaluate is reduced by a factor of 2 compared to the brute force method:

$$\sum_{A_1,\ldots,A_4} P(A_1, \ldots, A_4, A_5 = 1, A_6 = 1) = \mu_0 \mu_5 \mu_6 \mu_{56} \cdot$$
$$\cdot \sum_{A_4} (\mu_4 \mu_{46})^{G(A_4=1)} (\sum_{A_3} (\mu_3 \mu_{35})^{G(A_3=1)} \mu_{34}^{G(A_3=A_4=1)}$$
$$\cdot (\sum_{A_2} \mu_2^{G(A_2=1)} \mu_{23}^{G(A_2=A_3=1)} (\sum_{A_1} \mu_1^{G(A_1=1)})))$$

The distributive law thus allows for a more time-efficient implementation of the IS procedure.

### 3.4.2 Clique Tree Ideas

Another way to speed-up the IS algorithm is based on the decomposability of the probability distribution with respect to the graph of the model. A detailed treatment of such ideas can be found in [Pearl, 1988, Jirousek and Preucil, 1995, Malvestuto, 1992].

We first create a *chordal* graph $H'$ from $H$. To enforce chordality we use a graph triangulation algorithm [Pearl, 1988]. For the chordal graph, the joint probability distribution on the variables corresponding to its vertices can be decomposed into the product of the probability distributions on the maximal cliques of the graph divided over the product of the probability distributions on the clique intersections. The maximal cliques of the graph are placed into a *join tree* (or, in general, a forest) that shows how cliques interact with one another. The join or clique tree for the example in Figure 1 is given in Figure 2.

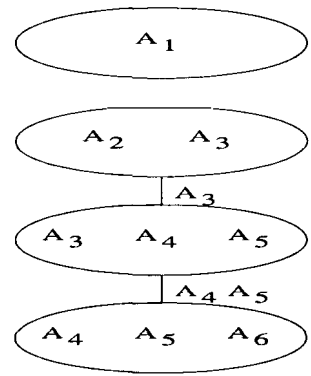

Figure 2: Clique forest corresponding to the problem in Figure 1. Cliques and their intersections are shown.

Thus, the original problem is decomposed into smaller problems corresponding to the cliques of the triangulated graph $H'$. Each smaller problem can be solved using IS, while distributions corresponding to the intersections can be found by summing out the corresponding distributions on the cliques. As we noted above, the time complexity of the IS algorithm grows exponentially with the size of the query. Thus, an algorithm that solves a number of smaller problems instead of solving a single large one may be considerably more efficient.

Note that although the bucket elimination and the clique tree approaches are similar in flavor, they are different. Bucket elimination uses the knowledge of the structure to perform the main summation more efficiently, and the rest of the IS algorithm remains unchanged, including the necessity to cycle over all constraints in each iteration. In the clique tree method we essentially use the fact that the maxent distribution defines an MRF and, thus, can equivalently be estimated as a product of functions corresponding to cliques of the graph of the model. We use regular IS algorithm to estimate the functions corresponding to cliques and return the product of clique distributions as a result.

Table 1: Characteristics of the Data Sets. $k$ is the number of attributes, $n$ is the number of records, $N_{1's}$ is the number of 1's in the data, $E(N_{1's}) = N_{1's}/n$, $Std(N_{1's})$ is the standard deviation of the number of 1's in the record, $Max(N_{1's})$ is the maximum number of 1's in the record.

|  | $k$ | $n$ | $N_{1's}$ | $E(N_{1's})$ | $Std(N_{1's})$ | $Max(N_{1's})$ |
|---|---|---|---|---|---|---|
| MS Web Data Set | 294 | 32711 | 98654 | 3 | 2.5 | 35 |
| Retail | 52 | 54887 | 224580 | 4.09 | 3.98 | 44 |

## 4 Empirical Results

### 4.1 Conjunctive Queries

We ran experiments on the two datasets: "The Microsoft Anonymous Web" dataset (publicly available at the UCI KDD archive) and a large proprietary dataset of consumer retail transactions. Both datasets contained binary transaction data. Before learning the models we analyzed the structure of the data and the itemsets that can be derived from it. Parameters of the datasets are given in the Table 1. The retail dataset is much more dense than the Microsoft Web Data. For more dense data sets, the larger itemsets will be more frequent, and the resulting graphs of the model will also be more dense.

We empirically evaluated (1) the independence model, (2) the Chow Liu tree model, and (3) the maxent model using the brute force, bucket elimination and clique tree methods. All experiments were performed on a Pentium III, 450 MHz machine with 128 Mb of memory. We generated 500 random queries for query sizes of 4, 6, and 8, and evaluated different models with respect to the average memory, online time, and error per Equation 1.

To select a query we first fixed the number of its variables $n_Q = 4$, 6 or 8. Then we picked $n_Q$ attributes according to the probability of the attribute taking a value of "1" and generated a value for each selected attribute according to its univariate probability distribution. Note, that negative values for the attributes are more likely in the sparse data than positive ones. Thus, generated queries typically had at most one positively initialized attribute.

The plots in Figure 3 show the dependence of the average relative error on the memory requirements for the model (or the model complexity, e.g., BIC) for the Microsoft Web data. The independence model using the least memory is the most inaccurate one. The Chow-Liu tree model exhibits intermediate performance between the independence and the maxent models.

Note that all the maxent models, i.e., brute force, bucket elimination and clique tree, have the same average error for a fixed query length since they es-timate the same product form of the distribution. Thus, on this figure we only report results for a single "maximum entropy model" (since all 3 produce the same estimates, but using different computational methods). Circles that show results for the maxent model correspond to various values of the threshold $T$ that was used to define itemsets. $T$ took values $15, 30, 50, 60, 100$ and $200$. The higher the value of $T$, the less information is supplied to the model and the less accurate the results are. Thus, the leftmost circle corresponds to $T = 200$ and the rightmost to $T = 15$.

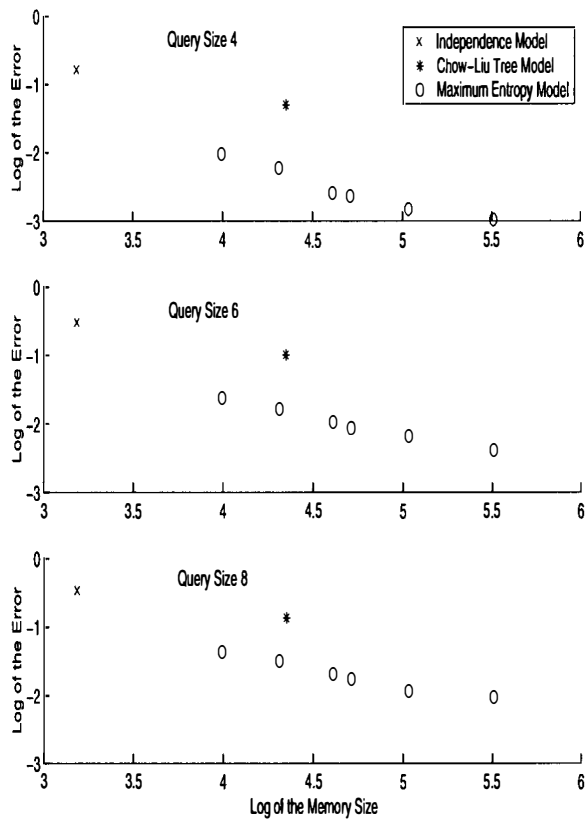

Figure 3: Average relative error on the 500 random queries as a function of model complexity for the Microsoft Web data. The $X$-axis reflects the number of parameters in the models.

The maxent model outperforms in terms of accuracy the tree model even when the amount of memory for the maxent model is lower. This assertion holds true for all the query sizes.

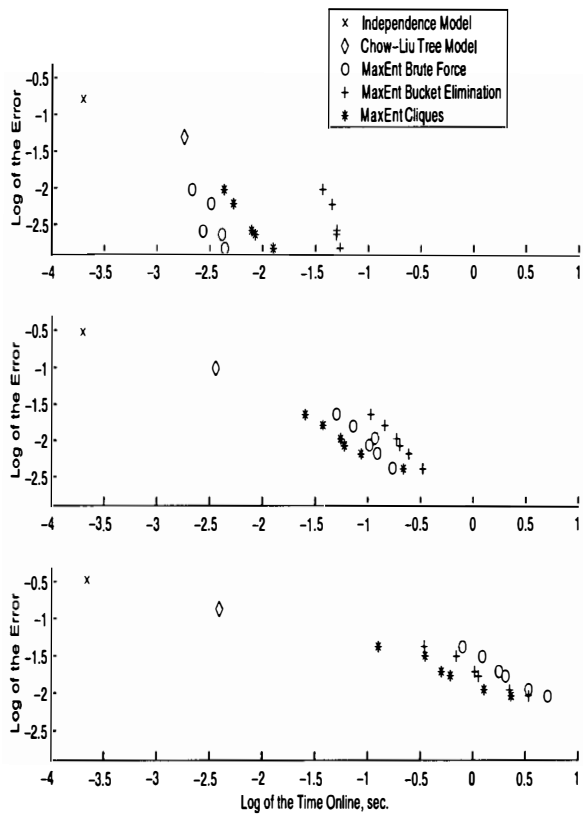

Figure 4: Average relative error on the 500 random queries as a function of the online time for the Microsoft Web data.

We also measured the average online time taken by various models to generate the estimated query count. Figure 4 illustrates how the error depends on the online time for all the models and query sizes 4, 6, and 8. Among the various models the independence model is the fastest but the least accurate of all. The Chow-Liu model fills in the large gap between the independence model and the cluster of the maxent models. On all three plots the diamond corresponding to the Chow-Liu model has nearly the same $x$-coordinate, showing a very slow increase in processing time as query size grows. The quality of the tree model is not as good as for the maxent models. Recall that the amount of information supplied to the Chow-Liu model is comparable to, and for some of the threshold values is even greater than, that for the maxent models.

The brute force, clique tree and bucket elimination maxent models have smaller errors than the other models but it takes them longer to produce the estimates. The error for all three types of maxent models is the same, and so is the $y$-coordinate for all points corresponding to the same threshold $T$; the only difference is in the online time. The brute force version is the fastest on the queries of size 4. This is not surprising as both the clique tree and the bucket elimination methods have an overhead that dominates for short queries. The clique tree method becomes the fastest for query sizes 6 and 8.

As the threshold $T$ decreases (the corresponding points on the plots go from left to right, top to bottom) and the number of itemsets (or memory size) increases, we see that the difference in the online time between the maxent algorithms that employ the graph structure and the brute force maxent decreases. We attribute this to the fact that when the number of itemsets increases, so does the average density of the graph of the model and the average induced width.

The performance of various models relative to one another on the *retail* data was qualititatively the same as for the Web data with the maxent model being again the most accurate but also the most computationally expensive. However, due to the fact that the retail data are much more dense, the memory requirements and the online running times were generally much higher than for the Web data.

### 4.2 Arbitrary Boolean Queries

It is straightforward to generalize the maxent approach to handle arbitrary Boolean queries. For a given arbitrary (not necessarily conjunctive) query we first estimate the maxent distribution on the query variables, then transform the query to disjunctive normal form and evaluate the distribution on the disjuncts. This approach is worst-case exponential in the query size.

We have run experiments on arbitrary Boolean queries that we generated according to the algorithm described above for conjunctive queries. The only difference is that the connective between two attributes was selected as either a disjunction or a conjunction by flipping a fair coin. Table 2 compares results on arbitrary and purely conjunctive queries ($n_Q$ is the query length, $t_P$, $C_t$ and $e_P$ are the average online time, query count and error across 200 runs of the algorithms). The maxent models again enjoy a distinct advantage in accuracy over the independence models.

## 5 Conclusions and Extensions

We have shown that (a) probabilistic models in general, and (b) the MRF/maxent approach in particular, provide a useful general framework for approximate query answering on large sparse binary datasets. We have empirically analyzed the relative performance of various probabilistic models for this problem and showed that given sufficient information about the data in the form of itemsets, the MRF/maxent approach is the most accurate of all the models. We also showed how bucket elimination and clique tree ideas can be employed for speeding up the learning of these models.

Table 2: Comparison of the Models on Arbitrary Boolean and Purely Conjunctive Queries.

| $n_Q$ | Conjunctive | | | | Arbitrary | | | |
|---|---|---|---|---|---|---|---|---|
| | $t_P$ | $C_t$ | $e_P$ | | $t_P$ | $C_t$ | $e_P$ | |
| | | | Indep. | Maxent | | | Indep. | Maxent |
| 4 | 0.052 | 14000 | 0.163 | 0.0021 | 0.059 | 25000 | 0.0066 | $8.2 \cdot 10^{-5}$ |
| 6 | 0.248 | 9200 | 0.304 | 0.0067 | 0.258 | 28000 | 0.0135 | $2.8 \cdot 10^{-4}$ |
| 8 | 2.036 | 6700 | 0.342 | 0.0112 | 2.525 | 29000 | 0.0319 | $6 \cdot 10^{-3}$ |

The work described in this paper allows for several possible extensions. For arbitrary Boolean queries one can in principle incorporate query structure directly into the IS algorithm or into bucket elimination. Another interesting problem is the issue of how threshold $T$ should be selected in practice, since the model complexity depends directly on $T$. Finally, there are several important open questions involving modeling of the query distribution: how should the query model be chosen? can it be learned from online user data? if it is known a priori, can it be profitably used in generating the approximate probability model, e.g., can one spend more resources on modeling parts of the data which have high probability of being queried?

### Acknowledgements

The research described in this paper was supported in part by NSF CAREER award IRI-9703120.